\documentclass[letterpaper, 10 pt, conference]{ieeeconf}  

\IEEEoverridecommandlockouts                              

\usepackage{cite}
\usepackage{booktabs}
\usepackage{amsmath,amssymb,amsfonts}
\usepackage{algorithmic}
\usepackage{graphicx}
\usepackage{textcomp}
\usepackage{xcolor}
\usepackage[table]{xcolor} 

\title{\LARGE \bf
UniSLAD: A Unified Framework for Structural and Logical Industrial Visual Anomaly Detection
}

\author{Changyi Li, Chao Yang, Yu Xiao, Kari Tammi}

\begin{document}

\maketitle

\begin{abstract}

Visual anomaly detection is a fundamental task in industrial automation. While existing approaches have achieved notable progress in identifying structural defects, the detection of logical anomalies remains relatively underexplored. In practice, structural and logical anomalies frequently co-occur in industrial workflows. Therefore, a solution capable of detecting both structural and logical anomalies is crucial for advancing comprehensive anomaly detection research. To address this limitation, we propose a unified framework, termed UniSLAD, which jointly addresses logical and structural anomalies without additional training, enabling a practical solution for dynamic industrial environments. First, we introduce a dual-feature extractor that synergistically integrates a Convolutional Neural Network (CNN) backbone for local texture perception with a Transformer backbone for global contextual reasoning, yielding richer and more comprehensive representations. Building on this foundation, we design dual-granularity feature representation modules. At the patch level, memory banks enhanced by the Mahalanobis Transform (MT) preserve representative features and support more discriminative anomaly scoring. At the image level, distribution maps are aggregated using Lower–Upper Mean (LUM) and Power Mean Pooling (PMP), yielding a more robust global representation than conventional average pooling. Extensive experiments on the two industrial benchmarks demonstrate that UniSLAD achieves competitive performance in comprehensive anomaly detection, achieving 99.4\% and 93.1\%, respectively. Furthermore, ablation studies verify the individual contributions and effectiveness of each proposed component. 

\end{abstract}

\begin{keywords}
anomaly detection, logical anomaly, structural anomaly, transformer, convolutional neural network
\end{keywords}

\section{INTRODUCTION}

\begin{figure}[t]
    \centering
    \includegraphics[width=0.43\textwidth]{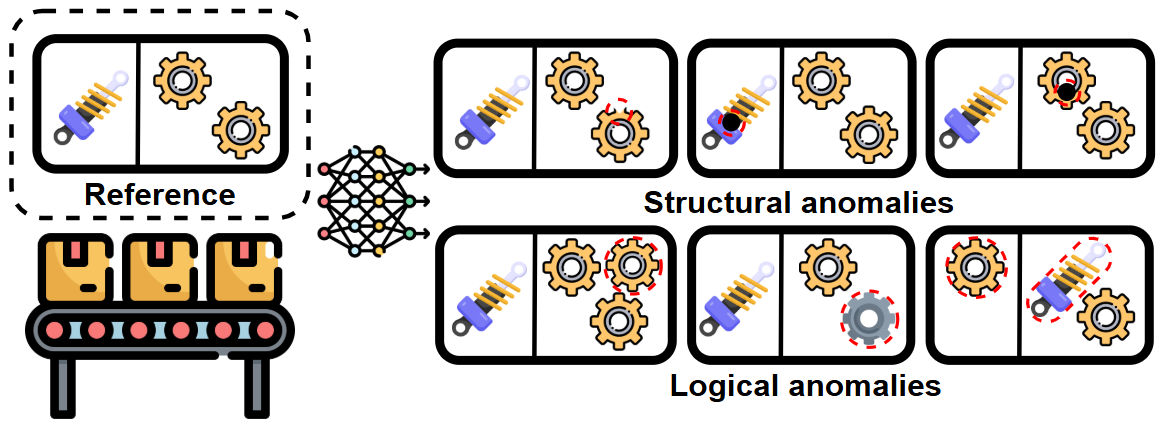} 
    \caption{Structural and logical anomalies within industrial automation.}
    \label{fig:1}
\end{figure}

Visual anomaly detection has become a cornerstone of industrial automation, enabling automated inspection and quality assurance. Leveraging advances in computer vision and deep learning, modern systems~\cite{zavrtanik2021draem, deng2022anomaly, 9879738, li2025s2tkd} have demonstrated strong capabilities in identifying localized structural anomalies, such as cracks, corrosion, scratches, or contamination, that directly affect component integrity. In parallel, logical anomalies, which involve contextual or relational inconsistencies such as incorrect counts, wrong types, or misplaced components, are also prevalent in real-world industrial settings. Importantly, structural and logical anomalies frequently co-exist, making their joint detection essential. As illustrated in Fig.~\ref{fig:1}, in logistics material kitting, it is not sufficient for individual components to be defect-free; the correct number, type, and model of items must also be present in each kit. Existing anomaly detection methods often focus on limited visual cues or feature granularities, making it difficult to simultaneously capture both structural defects and logical inconsistencies. These challenges underscore the need for a unified framework that jointly models structural and logical anomalies.

Methods that are effective for structural anomaly detection remain largely ineffective in detecting more complex global logical anomalies~\cite{zavrtanik2021draem, deng2022anomaly, 9879738, li2025s2tkd}. This gap arises because structural anomalies are typically localized and can be captured using patch-level features, whereas logical anomalies demand a holistic, global-level understanding. Recent studies~\cite{zhang2024contextual,batzner2024efficientad,yao2023learning} have explored dual-branch or hybrid anomaly detection frameworks based on Convolutional Neural Network (CNN) backbones, aiming to capture both local and global anomaly patterns. For instance, DSKD~\cite{zhang2024contextual} employs a dual-student knowledge distillation framework, where one student focuses on patch-level structural cues while the other targets image-level contextual relations. However, both students inherit the feature granularity of the same CNN-based teacher backbone, which is inherently biased toward local features and thus limits their ability to integrate truly heterogeneous representations. EfficientAD~\cite{batzner2024efficientad} takes a different approach by combining a student–teacher branch for structural anomalies with an autoencoder branch for global analysis. While this design improves efficiency, it still relies on single-scale feature spaces within each branch, and reconstruction-based autoencoder risk restoring anomalous regions as normal, leading to missed detections.

These limitations highlight the need for frameworks that can natively integrate complementary feature granularities, rather than merely extending a single backbone or relying on reconstruction. A particularly promising direction is the combination of CNN-based and Transformer-based architectures. CNNs (e.g., ResNet~\cite{he2016deep}) excel at capturing local textures and fine-grained structural details through convolutional kernels, while Transformers leverage self-attention to model long-range dependencies and contextual relationships across all patches. This complementarity suggests that jointly leveraging CNNs and Transformers could enable more robust detection of both localized structural anomalies and global logical inconsistencies. Nevertheless, how to effectively fuse these heterogeneous feature representations into a unified anomaly detection framework remains an open and underexplored challenge.

To tackle this challenge, we propose UniSLAD (Unified Structural and Logical Anomaly Detection), a unified framework designed to integrate different features and detect both structural and logical anomalies. 
First, we introduce a dual-feature extractor that synergistically integrates ResNet for local texture perception and Vision Transformers (ViTs)~\cite{dosovitskiy2020image} for global understanding. Second, we design dual-granularity feature representation modules to enhance anomaly discrimination. At the patch level, all features are projected into a decorrelated space via the Mahalanobis Transform (MT), allowing representative features to be stored in memory banks for fine-grained anomaly scoring. At the image level, we introduce a novel aggregation strategy that combines Lower–Upper Mean (LUM) with Power Mean Pooling (PMP) to yield more expressive and robust global representations, constructing representative feature distributions.
Finally, an integrated anomaly detector fuses evidence from both patch-level memory banks and image-level distribution maps, ensuring consistent detection of structural defects and logical inconsistencies.
Through extensive experiments on the MVTec AD and MVTec LOCO AD datasets, we demonstrate that UniSLAD delivers highly competitive performance with AUROCs of 99.4\% and 93.1\%, respectively. Our ablation studies further validate the effectiveness of each proposed component, confirming the superiority of our unified design in real-world industrial settings where structural and logical anomalies co-exist.

This contribution of this work can be summarized as:
\begin{itemize}
    \item Propose \textbf{UniSLAD}, a unified framework that effectively fuses heterogeneous features to detect both structural and logical anomalies.
    \item Design a \textbf{Dual-Feature Extractor} that integrates ResNet and ViT backbones to extract both local and global features.
    \item Introduce the \textbf{Mahalanobis Transform} and a novel \textbf{LUM–PMP aggregation strategy}, enhancing feature representation across multiple granularities.   
    \item Achieve competitive performance on MVTec AD and LOCO AD, with ablation studies validating the contribution of each component.
\end{itemize}

\section{Related Work}
While significant progress has been made in structural anomaly detection, with methods such as DRAEM~\cite{zavrtanik2021draem}, RD4AD~\cite{deng2022anomaly}, and PatchCore~\cite{9879738} demonstrating promising performance, these approaches frequently exhibit limited effectiveness in identifying logical anomalies. This shortcoming has established logical anomaly detection as a critically underserved and emerging area of research. 
Yao et al.~\cite{yao2023visual} proposed DADF, a framework leveraging a dual-attention Transformer and normalizing flow for feature reconstruction and discriminative likelihood estimation in unsupervised anomaly detection. Yang et al.~\cite{yang2023slsg} introduced SLSG, which employs self-attentive graph convolution and simulated anomalies to model normal patterns and improve logical anomaly detection through graph-based relationship learning. Yao et al.~\cite{yao2023learning} presented GLCF, a two-branch Transformer architecture designed to jointly capture structural and logical anomalies by integrating local and global features via a semantic bottleneck. However, these methods solely employ either CNNs or Transformers as feature extractors, which limits their ability to efficiently capture both local features and global context. To overcome this limitation, we propose a dual feature extractor that combines the strengths of both architectures, enabling the model to simultaneously identify structural and logical anomalies.

In addition, Guo et al.~\cite{guo2023template} proposed a Template-guided Hierarchical Feature Restoration method for multi-scale anomaly detection, which uses bottleneck compression to retain essential normal patterns and template-guided compensation to restore distorted features using similar normal samples. Anomalies are identified via cosine distance between original and restored features. EfficientAD~\cite{batzner2024efficientad} employed a student-teacher framework where the student is trained to mimic normal features from the teacher; anomalies are detected based on prediction deviations. It incorporates a constrained imitation loss and uses an autoencoder to improve global anomaly reasoning. Although these methods have achieved notable performance, they primarily rely on either patch-level or image-level features alone, leading to a significant performance gap in detecting both structural and logical anomalies. To address this limitation, we propose a multi-granularity feature fusion approach that enables robust performance across both types of anomalies.
In summary, to address the limitations of existing methods, we design a training-free unified framework that incorporates a dual feature extractor and multi-granularity feature representation, enabling efficient and simultaneous detection of both structural and logical anomalies.

\section{Method}

\begin{figure*}[t]
    \centering
    \includegraphics[width=0.8\textwidth]{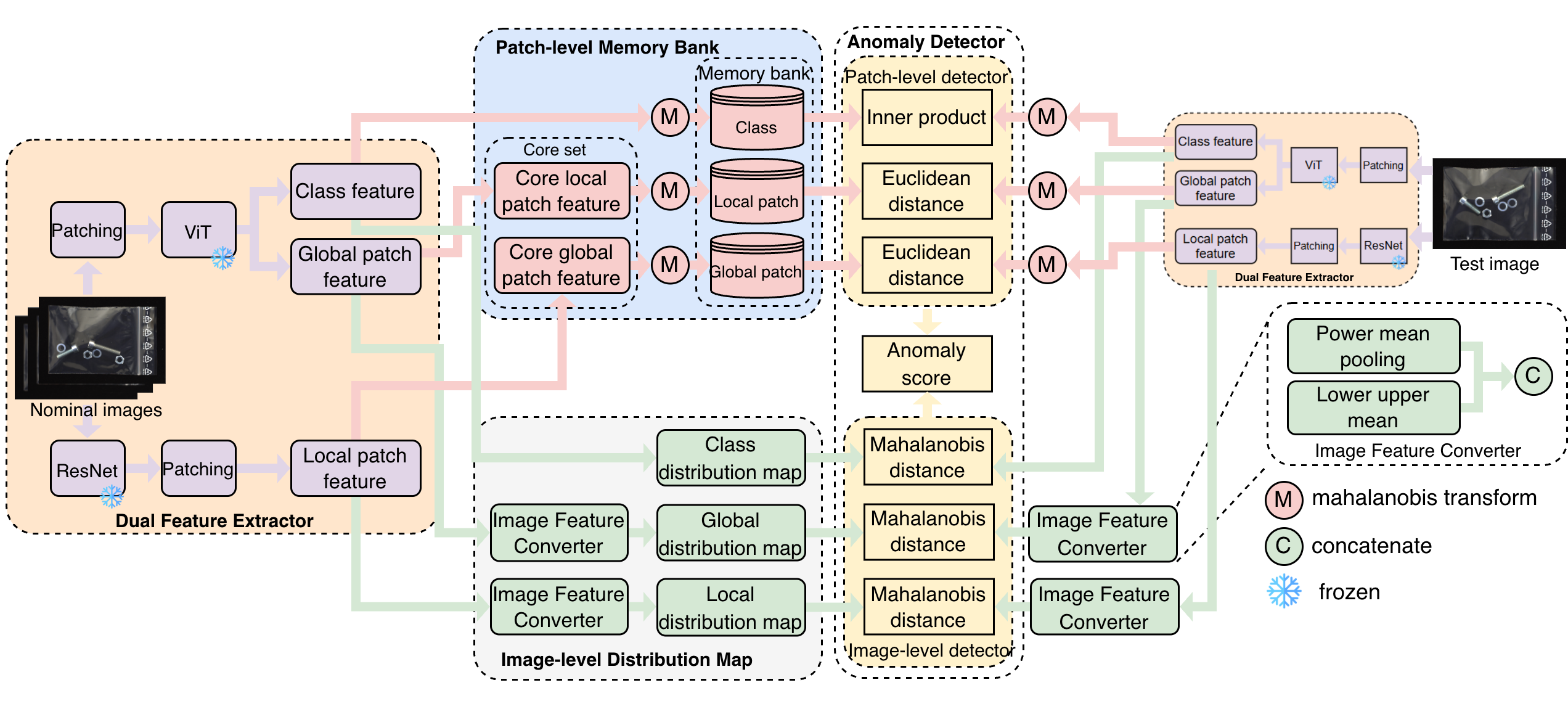} 
    \caption{Overall pipeline of the proposed UniSLAD. Nominal images (left side) are first fed into the Dual-Feature Extractor to obtain local, global, and class features. These features are used to construct patch-level memory banks and image-level distribution maps as nominal references. During test, the test image (right side) is processed by the same feature extractor, and its features are compared with the nominal memory banks and distribution maps using multiple similarity measures to compute anomaly scores.}
    \label{fig:framework}
\end{figure*}
Fig.~\ref{fig:framework} illustrates the proposed UniSLAD framework for both structural and logical anomaly detection. It consists of four main components. 
First, a \textbf{Dual-Feature Extractor} integrates a ResNet backbone for local texture-sensitive features with a ViT backbone for global contextual representations, providing complementary feature embeddings. Next, the \textbf{Patch-level Memory Bank} selects representative patch features using coreset sampling and applies the Mahalanobis Transform to construct compact memory banks. In parallel, the \textbf{Image-level Distribution Map} aggregates patch features into holistic image features through PMP and LUM, yielding robust distributional representations. 
Finally, the \textbf{Anomaly Detector} compares test features with the nominal memory banks at the patch level and with the distribution maps at the image level, enabling the detection of structural and logical anomalies. The following subsections describe each module in detail.

\subsection{Dual-Feature Extractor}\label{DFE}
To effectively capture both structural and logical anomalies, we employ the \textbf{Dual-Feature Extractor} that integrates complementary representations. A ResNet backbone is used to extract local, texture-sensitive features that are well-suited for identifying fine-grained structural irregularities, while a ViT backbone captures long-range dependencies and global contextual information, which are essential for detecting logical inconsistencies.

We denote the nominal image set as $X_{\text{nominal}}$ and the testing set as $X_{\text{test}}$, where each image $x_i \in X_{\text{nominal}} \cup X_{\text{test}}$ satisfies $x_i \in \mathbb{R}^{H \times W \times C}$, with $H$, $W$, and $C$ representing the image height, width, and number of channels, respectively.

\textbf{CNN-based local feature extractor: }
A ResNet backbone, denoted as $\varphi$ and pre-trained on ImageNet~\cite{5206848}, is employed to extract intermediate feature maps across different blocks. 
Formally, the feature map extracted from layer $j$ for input image $x_i$ is denoted as $\varphi_j(x_i)$.
 
The feature representation $\varphi_j$ is formed by all residual layers from the first up to the $j$-th layer.
For an input image $x_i$, the feature vector at spatial location $(h,w)$ in $\varphi_j(x_i)$ is denoted by $\varphi^{h,w}_j(x_i)$, 
where $h$ and $w$ index the row and column coordinates of the feature map. 
This vector is interpreted as a patch-level descriptor, serving as the fundamental unit for subsequent local feature extraction. 
To incorporate local spatial context, we define the neighborhood of $(h,w)$ with a window size $p$ as

\begin{equation}
\begin{split}
N^{h,w}_p = &\left\{ (h', w') \mid h' \in \left[h-\left\lfloor p/2 \right\rfloor, \dots, h + \left\lfloor p/2 \right\rfloor \right] \right., \\
&\quad w' \in \left[w - \left\lfloor p/2 \right\rfloor, \dots, w + \left\lfloor p/2 \right\rfloor \right] \},
\end{split}
\end{equation}
\noindent where $(h', w')$ indexes spatial positions within the defined neighborhood.

To fuse the neighborhoods of the feature vector and increase the receptive field of our shallow features, we concatenate the neighborhoods to form the locally aware patch feature, as follows:
\begin{equation}
f^{(h,w),l}_i = f_{\text{agg}}\left( \varphi^{h',w'}_j(x_i) \mid \left( h', w' \in N^{h,w}_p \right) \right).
\end{equation}

$f^{(h,w),l}_i$ represents the final local patch feature of $x_i$ at spatial location $(h,w)$, obtained by aggregating the features within its neighborhood $N^{h,w}_p$. $f_{\text{agg}}$ denotes the locally aware adaptive average pooling aggregator~\cite{9879738}. 

\textbf{Transformer-based global feature extractor: }
Complementing the CNN backbone, we employ ViT~\cite{dosovitskiy2020image}, denoted as $\mathcal{T}$, which is pre-trained on ImageNet~\cite{5206848}. 
In ViT-based feature extractor, the image $x_i$ is partitioned into $N = HW/P^2$ non-overlapping patches of size $(P,P)$, denoted as $\{x^{p_1}_i, x^{p_2}_i, \dots, x^{p_N}_i\}, x^{p_k}_i \in \mathbb{R}^{P \times P \times C}$.
Each patch $x^{p_k}_i$ is flattened into a vector $p_i^k = \mathrm{Flatten}(x^{p_k}_i) \in \mathbb{R}^{P^2 \cdot C}$, linearly projected using a matrix $\mathbf{E} \in \mathbb{R}^{(P^2 \cdot C) \times D}$, and combined with a positional encoding $e^{pos} \in \mathbb{R}^{(N+1) \times D}$. This process yields the embedding $\tilde{z}_i \in \mathbb{R}^{(N+1) \times D}$:  

\begin{equation}
\tilde{z}_i = [x_{class}; p_i^1\mathbf{E}; p_i^2\mathbf{E};\dots; p_i^N\mathbf{E}] + {e^{pos}},
\end{equation}
\noindent where $x_{class}$ denotes a [CLS] token prepended to the sequence. The resulting embeddings are processed by $\mathcal{T}$, which exploits self-attention to capture dependencies among patches. The [CLS] token progressively aggregates global contextual information, thereby enabling the ViT to simultaneously extract localized features from individual patches and a global representation of the entire image. Formally, the output of the first $l$ layers of $\mathcal{T}$ is expressed as  
\begin{equation}
[f_i^{l,c}, f_i^{l,g}] = \mathcal{T}_l(\tilde{z}_i)
\end{equation}
\noindent where $\mathcal{T}_l$ denotes the truncated ViT up to layer $l$, $f_i^{l,c}$ represents the class feature, and $f_i^{l,g}$ corresponds to the global patch feature representations.

\subsection{Patch-level Memory Bank}\label{PAD}
To enable efficient and discriminative representation learning, 
the dense set of extracted patch features is transformed into a compact and structured form. The \textbf{Patch-level Memory Bank} consists of two components. First, representative subsets of local and global patch features are selected using the greedy coreset algorithm~\cite{agarwal2005geometric}, 
which reduces redundancy while preserving the diversity of nominal patterns. Second, to mitigate correlations among selected features that may impair the accuracy of nearest-neighbor matching, we introduce the Mahalanobis Transform. This transformation decorrelates the feature space, thereby enhancing the discriminative capacity of the memory banks and improving anomaly detection performance.  

\textbf{Coreset: }
Given the large number of extracted patch features, we employ the greedy coreset algorithm~\cite{agarwal2005geometric} to construct compact subsets of representative features. Formally, for a feature set $F \in \{F_L, F_G\}$, the selected subset is obtained
\begin{equation}
F_p = \text{CoreSet}(F),
\end{equation}
where $\text{CoreSet}(\cdot)$ represents the greedy coreset algorithm, $F_L = \{ f^{(h,w),l}_i \}$ denotes the collection of local patch features, 
and $F_G = \{ f^{l,g}_i \}$ denotes the collection of global patch features,
and $F_{lp}$ and $F_{gp}$ are their corresponding selected subsets.

\textbf{Mahalanobis Transform: }
While coreset selection reduces redundancy, the resulting features may still exhibit correlations that negatively affect similarity search during anomaly detection. To address this issue, we apply the Mahalanobis Transform, which projects features into a decorrelated space and improves their representational capacity. 
The procedure consists of four main steps: (1) covariance matrix computation, (2) covariance matrix regularization and inversion, (3) eigenvalue decomposition for whitening, and (4) feature normalization.  

We take local patch features as an example. First, the covariance matrix of all features is computed as
\begin{equation}
\Sigma = \frac{1}{n-1} \sum_{c=1}^n (f^p_c - \bar{f}^p)(f^p_c - \bar{f}^p)^\top,
\end{equation}
where $\bar{f}^p = \frac{1}{n} \sum_{c=1}^n f^p_c$, $f^p_c \in F_{lp}$, is the mean feature vector, and $n$ is the total number of local patch features in $F_{lp}$. 
Next, the inverse covariance matrix is obtained and decomposed as
\begin{equation}
\Sigma^{-1} = U \Lambda U^\top,
\end{equation}
where $U$ contains the eigenvectors and $\Lambda$ is a diagonal matrix of eigenvalues. The inverse square root matrix is constructed as
\begin{equation}
L_2 = U \sqrt{\max(\Lambda, 0)} U^\top,
\end{equation}
where negative eigenvalues are truncated to zero and square roots are applied elementwise. Each feature $f^p_c$ is then mean-centered and transformed:
\begin{equation}
o_c = (f^p_c - \bar{f}^p) L_2,
\end{equation}
followed by row-wise $\ell_2$ normalization:
\begin{equation}
\hat{o}_c = \frac{o_c}{\|o_c\|_2}.
\end{equation}

The transformed features $\hat{o}_c$ constitute the local patch feature memory bank:
\begin{equation}
M_{lp} = \bigcup_{c=1}^n \hat{o}_c, \quad \hat{o}_c = MT(f^p_c), \; f^p_c \in F_{lp},
\end{equation}
where $M_{lp}$ denotes the memory bank of local patch features, 
and $MT(\cdot)$ indicates that the Mahalanobis Transform is applied to each feature. The same procedure is applied to global patch features $F_{gp}$ and class patch features $F_{C}=\{f_i^{l,c}\}$, yielding the memory bank $M_{gp}$ of global patch features and the memory bank $M_{c}$ of class features, respectively.

\subsection{Image-level Distribution Map}\label{IAD}
The \textbf{Image-level Distribution Map} is designed to provide a compact representation of each image for distribution-based anomaly detection. It computes holistic image features by aggregating patch features within each image. 
For each category, the statistical measures (mean and covariance) of the nominal images are then estimated, 
forming the nominal feature distribution used for anomaly detection.

Conventional approaches~\cite{10647438, liu2024image} obtain image features by simply averaging all patch features. 
Although straightforward, this strategy often fails to capture informative high-response regions, resulting in limited discriminative power. To address this limitation, we propose a more expressive aggregation scheme that fuses patch features 
into a richer image-level representation, thereby improving the accuracy of nominal distribution estimation and enhancing the detection of anomalies.  
Specifically, we combine Power Mean Pooling (PMP) and Lower-Upper Mean (LUM).  
PMP generalizes average pooling by introducing a tunable parameter $q$ that controls the balance between 
average and max pooling. Setting $q>1$ emphasizes highly activated regions in the feature map, 
making the representation more sensitive to localized anomalies. 
Formally, for local patch features, we define
\begin{equation}
f_{\mathrm{pmp}}^l = \left(\frac{1}{K^2} \sum_{h,w=1}^{K} \big(f^{(h,w),l}_i\big)^q \right)^{\tfrac{1}{q}},
\end{equation}
where $f_{\mathrm{pmp}}^l$ is the pooled local feature vector, $q$ is the power parameter, and $K^2$ denotes the number of local patch features.

While PMP emphasizes salient regions, it is sensitive to outliers and noisy patches. 
To enhance robustness, we incorporate LUM pooling, which discards a fixed proportion of the largest and the smallest feature responses before averaging the remainder. 
This suppresses extreme values and yields more stable features. Specifically, all local patch features $f^l_i$ of the image $x_i$ are sorted to obtain $v^l_i$, the image feature using LUM can be defined as:
\begin{equation}
\begin{split}
f_{\mathrm{lum}}^l &= \frac{1}{K^2 - 2t} \sum_{r=t+1}^{K^2-t} v^{r,l}_i,
\end{split}
\end{equation}
where $f_{\mathrm{lum}}^l$ represents the local feature derived from LUM pooling, $v^{r,l}_i$ represents the $r$-th patch feature in the ordered sequence $v^l_i$. 
The number of data points excluded from each end of the sequence is given by
$t = floor(\alpha * K^2)$, 
where $floor$ denotes the floor function to ensure that $t$ is an integer, $\alpha$ is the trimming proportion.

The final image-level feature is obtained by concatenating the outputs of PMP and LUM:
\begin{equation}
f_{\mathrm{img}}^l = \mathrm{concat}\big(f_{\mathrm{pmp}}^l, f_{\mathrm{lum}}^l\big),
\end{equation}
where $f_{\mathrm{img}}^l$ denotes the aggregated local image feature, $\mathrm{concat}$ donates the concatenation operation. 
Similarly, the aggregated global image feature $f_{\mathrm{img}}^g$ is obtained by applying the same procedure. 
In addition, the class feature $f_i^{l,c}$ from the ViT-based extractor serves as another image-level representation. Together with the aggregated local and global features $f_{\mathrm{img}}^l$ and $f_{\mathrm{img}}^g$, 
these three representations form the basis for image-level analysis.  
For the nominal image set $X_{\text{nominal}}$, we compute the mean and covariance of each representation to characterize the nominal distributions:  
\begin{equation}
\mathcal{D}_{\mathrm{img}} = 
\Big\{ (\mu_c^l, \Sigma_c^l), \; (\mu_c^g, \Sigma_c^g), \; (\mu_c^{cls}, \Sigma_c^{cls}) \;\Big|\; c \in \mathcal{C} \Big\},
\end{equation}
where $(\mu_c^l, \Sigma_c^l)$, $(\mu_c^g, \Sigma_c^g)$, and $(\mu_c^{cls}, \Sigma_c^{cls})$ 
denote the mean and covariance of $f_{\mathrm{img}}^l$, $f_{\mathrm{img}}^g$, and $f_i^{l,c}$, respectively. $\mathcal{C}$ denotes the set of categories, and the distribution map $\mathcal{D}_{\mathrm{img}}$ acts as the statistical baseline for image-level anomaly detection.

\subsection{Anomaly Detector}\label{fusion}
The \textbf{Anomaly Detector} integrates patch-level and image-level evidence to identify both structural and logical anomalies.  
During the testing phase, the test image in $X_{\text{test}}$ is first fed through the \textbf{Dual-Feature Extractor} to obtain the class feature $f^c_t$, global patch feature $f^{g,p}_t$, and local patch feature $f^{l,p}_t$. These representations are subsequently passed to the patch-level and image-level detectors, where anomaly scores are computed and fused into a final decision.

\textbf{Patch-level detector:}
After applying the Mahalanobis Transform, 
the global and local patch features are denoted as  $\hat{f}^{g,p}_t$, and $\hat{f}^{l,p}_t$.
The patch-level anomaly scores are computed as the minimum Euclidean distances via Facebook AI Similarity Search (FAISS) between these features and the corresponding memory banks:
\begin{equation}
S_p^l = \min_{m \in M_{lp}} \| \hat{f}^{l,p}_t - m \|_2, 
\quad 
S_p^g = \min_{m \in M_{gp}} \| \hat{f}^{g,p}_t - m \|_2,
\end{equation}
where $S_p^l$ and $S_p^g$ denote the local and global patch-level anomaly scores, respectively. To adaptively fuse them, the similarity between the class feature of the test image $f_t^c$ 
and the class features in the class memory bank $M_c$ is computed using FAISS’s inner product:
\begin{equation}
w_p = IP(f_m^c, f_t^c) = \sum_{i=1}^d f_m^{c,i} f_t^{c,i},
\label{similarity}
\end{equation}
where $f_m^c \in M_c$ and $d$ is the feature dimension.  
This similarity $w_p$ serves as a weight to combine the local and global scores:
\begin{equation}
S_p = (1-w_p) \cdot S_p^g + w_p \cdot S_p^l,
\label{patch anomaly score}
\end{equation}
where $S_p$ is the final patch-level anomaly score.

\textbf{Image-level detector:}
For Image-level anomaly detection, the global patch features $f^{g,p}_t$ and local patch features $f^{l,p}_t$ are processed through the Image Feature Converter to obtain the global image feature $\bar{f}^{g,p}_t$ and local image feature $\bar{f}^{l,p}_t$. The Mahalanobis distance~\cite{mahalanobis2018generalized} is used to measure the deviation 
of test features from the nominal distributions.  
For the local image feature $\bar{f}^{l,p}_t$, the anomaly score is computed as
\begin{equation}
S^l_i = \sqrt{(\bar{f}^{l,p}_t - \mu_c^l)^\top {\Sigma_c^l}^{-1} (\bar{f}^{l,p}_t - \mu_c^l)},
\label{MD}
\end{equation}
where $\mu_c^l$ and $\Sigma_c^l$ are the mean and covariance of the nominal local image features. 
Analogous scores are computed for the global image feature $\bar{f}^{g,p}_t$ and the class feature $f^c_t$. These are fused into the global image-level score $S_i^g$ following the same weighted fusion structure as Eq.~(\ref{patch anomaly score}), the weighting parameter is $w_{ig}$.
The overall image-level anomaly score is then computed as
\begin{equation}
S_i = (1-w_i) S^l_i + w_i S^g_i,
\label{image anomaly score}
\end{equation}
where $w_i$ is a weighting hyperparameter. Finally, patch-level and image-level scores are integrated to yield the overall anomaly score:
\begin{equation}
S = (1-w) S_i + w S_p,
\label{anomaly score}
\end{equation}
with $w$ balancing the contributions of the two levels.

\begin{table*}[htbp]
    \centering
    \caption{
    I-AUROC/P-AUROC(\%) results of different models on the MVTec AD dataset. The best results among all methods are marked in \textbf{BOLD}.
    Unavailable results are denoted with ‘-’. * indicates types that contain both structural and logical anomalies. }
    \label{tab:localization}
    {
    \resizebox{\textwidth}{!}{
    \begin{tabular}{ccccccccc>{\columncolor{gray!30}}c}
        \hline
        Type & DRAEM  & RD4AD & THFR & PatchCore & DADF & EfficientAD & GRAD& SLSG& UniSLAD \\
        \hline
        carpet      & 97.0/95.5  &98.9/98.9 & \textbf{99.8}/\textbf{99.2} & 98.7/99.0 & 98.1/98.9 & 99.4/- & 98.2/96.5 & 99.0/96.0 & 99.5/98.9\\
        grid        & 99.9/\textbf{99.7}  & \textbf{100.0}/99.3 & \textbf{100.0}/99.3& 98.2/98.7 & \textbf{100.0}/99.4 & \textbf{100.0}/- & \textbf{100.0}/97.2& \textbf{100.0}/98.5 & 99.9/98.8\\
        leather    & \textbf{100.0}/98.6 & \textbf{100.0}/99.4 & \textbf{100.0}/99.4&  \textbf{100.0}/99.3 & \textbf{100.0}/\textbf{99.6} & \textbf{100.0}/- & \textbf{100.0}/98.8 & \textbf{100.0}/99.5 & \textbf{100.0}/\textbf{99.6}\\
        tile        & 99.6/\textbf{99.2}  &99.3/95.6 & 99.3/95.5 & 98.7/95.6 & \textbf{100.0}/97.3 & 99.9/- & \textbf{100.0}/95.4 & \textbf{100.0}/98.6 & 99.1/97.7\\
        wood        & 99.1/96.4  &99.2/95.3 & 99.2/95.3 &  99.2/95.0 & 97.8/96.4 & 99.5/- & 98.3/87.2 & 99.6/\textbf{96.8} & \textbf{100.0}/96.2\\
        bottle      & 99.2/99.1  & \textbf{100.0}/98.7 & \textbf{100.0}/98.9 & \textbf{100.0}/98.6 & \textbf{100.0}/98.7 & \textbf{100.0}/- & \textbf{100.0}/96.5 & 99.4/99.1 & \textbf{100.0}/\textbf{99.3}\\
        \textbf{cable*}       & 91.8/94.7  & 95.0/97.4 & 99.2/\textbf{98.5} & 99.5/98.4 & 99.2/99.0 & 97.3/- & 99.3/98.4 & 98.3/97.4 & \textbf{99.7}/97.7\\
        \textbf{capsule*}     & 98.5/94.3  & 96.3/98.7 & 97.5/98.7 & 98.1/98.8 & 95.0/98.6 & 98.4/- & 96.4/97.1 & 95.5/95.9 & \textbf{98.7}/\textbf{99.0}\\
        hazelnut    & \textbf{100.0}/\textbf{99.7}  & 99.9/98.9 & \textbf{100.0}/99.2 & \textbf{100.0}/98.7 & 96.0/98.6 & 99.6/- & 98.1/96.6 & 99.5/97.8 & \textbf{100.0}/99.3\\
        metal\_nut   & 98.7/\textbf{99.5}  & \textbf{100.0}/97.3 & \textbf{100.0}/97.4 & \textbf{100.0}/98.4 & 99.7/97.6 & 99.0/- & \textbf{100.0}/93.7 & \textbf{100.0}/98.9 & 99.7/99.2\\
        pill       & 98.9/97.6  & 96.6/98.2 & 97.8/98.0 & 96.6/97.4 & 97.2/97.8 & 98.2/- & 95.7/98.1 & \textbf{99.2}/98.0 & 97.9/\textbf{99.0}\\
        screw      & 93.9/97.6 & 97.0/\textbf{99.6} & 97.1/99.5/ &  \textbf{98.1}/99.4 & 95.8/99.1 & 97.4/- & 96.0/99.2 & 89.1/97.3 & 97.9/98.3\\
        toothbrush  & \textbf{100.0}/98.1 & 99.5/\textbf{99.1} & \textbf{100.0}/99.2 &  \textbf{100.0}/98.7 & 97.2/98.2 & \textbf{100.0}/- & 99.7/98.0 & \textbf{100.0}/99.4 & \textbf{100.0}/98.9\\
        \textbf{transistor*}  &93.1/90.9  &96.7/92.5 & 99.7/95.9 &  \textbf{100.0}/96.3 & \textbf{100.0}/\textbf{98.6} & 99.7/- & \textbf{100.0}/97.8 & 97.3/92.5 & \textbf{100.0}/98.5\\
        zipper      & \textbf{100.0}/\textbf{98.8} & 98.5/98.2 & 97.7/98.7 & 99.4/98.5 & 98.5/98.7 & 98.4/- & 99.7/98.3 & \textbf{100.0}/97.1 & 99.0/\textbf{98.8}\\
        \hline
        average    &98.0/97.3 & 98.5/97.8 & 99.2/98.2 & 99.1/98.1 & 98.3/98.4 & 99.1/- & 98.7/96.8 & 98.5/97.5 & \textbf{99.4}/\textbf{98.6}\\
    \hline
    \end{tabular}}
    }
\end{table*}

\section{Experiments}
\subsection{Datasets}
We conducted comprehensive experiments on two widely used industrial anomaly detection datasets, including the MVTec LOCO AD dataset and MVTec AD dataset.

{\bf MVTec AD}~\cite{8954181} comprises 15 categories. Most anomalies in the dataset are structural; however, for the categories Cable, Capsule, and Transistor, the anomalies include both structural and logical anomalies. To support evaluation, pixel-level binary annotations are available for anomalous test images, enabling fine-grained assessment of defect localization.

{\bf MVTec LOCO AD}~\cite{10.1007/s11263-022-01578-9} is designed to evaluate unsupervised anomaly localization algorithms and includes both structural and logical anomalies. It consists of 3,644 images. Structural anomalies include defects such as scratches, dents, or contaminations, while logical anomalies involve violations of underlying constraints, such as objects being in invalid locations or missing entirely. Ground truth for structural and logical anomalies is provided as pixel-level binary masks.

\subsection{Evaluation Metrics} 
We use two widely used metrics for evaluation. I-AUROC assesses anomaly detection by measuring how well the model distinguishes normal from anomalous images, while P-AUROC evaluates anomaly localization by comparing the anomaly score map with pixel-level ground truth. Together, these metrics provide a comprehensive view of both detection and localization performance.

\subsection{Implementation Details}
We conducted all experiments using an NVIDIA A40 GPU with 40 GB of memory and an Intel(R) Xeon(R) Platinum 8480+ CPU with PyTorch. For the CNN-based feature extractor, we employ wide-resnet50 as the backbone. For the Transformer-based feature extractor, we use ViT-B/16 as the base model. For our method, all images were resized to $224 \times 224$ pixels. Additionally, when computing the patch-level anomaly score using Euclidean distances, the $k$-nearest was set to 1. $\alpha$ was set to 0.05. $q$ was set to 3. $K$ was set to 56. The window size $p$ was set to 3. 

\subsection{Comparisons with the State-of-the-art}
We selected a total of 14 state-of-the-art methods as baselines, including DREAM, RD4AD, THFR, PatchCore, DADF, EfficientAD, GRAD, SLSG, ComAD, SAM-LAD, LogSAD, LogicAD, GCAD, and SINBAD. Among these, some methods primarily focus on structural anomaly detection, such as DREAM, RD4AD, and PatchCore. The other methods were designed to address both structural and logical anomaly detection.

\textbf{MVTec AD: }
In this dataset, we use I-AUROC to compare the anomaly detection performance of various methods. Notably, the categories \textbf{Cable}, \textbf{Capsule}, and \textbf{Transistor} contain both structural and logical anomalies, making them more indicative of a model's ability to jointly detect the two anomaly types. By contrast, the remaining categories involve only structural anomalies and thus primarily reflect performance on structural defect detection. As shown in Table \ref{tab:localization}, our method achieves an overall I-AUROC of 99.4\% on this dataset, outperforming all competing methods. In particular, for the Cable, Capsule, and Transistor categories, our method attains the best results, with I-AUROC scores of 99.7\%, 98.7\%, and 100.0\%, respectively. These findings highlight the effectiveness of the proposed approach in addressing both structural and logical anomaly detection. Additionally, our model also outperforms other methods in P-AUROC, achieving 98.6\%.

\begin{table*}[htbp]
\centering
\caption{I-AUROC(\%) results (logical/structural/average) on MVTec LOCO AD dataset. $\dagger$ indicates training-free approaches. $*$ indicates methods that utilize additional annotations.}

\label{tab:auroc_mvtec_loco}
\begin{tabular}{l c c c c c c c}
\toprule
Method  & Breakfast Box & Juice Bottle & Pushpins & Screw Bag & Splicing Connectors & Average \\
\midrule

DRAEM~\cite{zavrtanik2021draem}  & 75.1/85.4/- & 97.8/90.8/- & 55.7/81.5/- & 56.2/85.0/- & 75.2/95.5/- & 72.0/87.6/79.8 \\

RD4AD~\cite{deng2022anomaly}  & 66.7/60.3/- & 93.6/95.2/- & 63.6/84.8/- &54.1/89.2/- & 75.3/95.9/- & 70.7/85.1/77.9 \\
PatchCore~\cite{9879738}$\dagger$ & 74.8/80.1/- & 93.9/98.5/- & 63.6/87.9/- & 57.8/92.0/- & 79.2/88.0/- & 74.0/89.3/81.7 \\
GCAD~\cite{10.1007/s11263-022-01578-9}  & 87.0/80.9/- & \textbf{100.0}/98.9/- & \textbf{97.5}/74.9/- & 56.0/70.5/- & 89.7/78.3/- & 86.0/80.7/83.4 \\
ComAD~\cite{liu2023component}  & 94.7/70.0/- &90.9/80.5/- & 89.0/93.8/- & 79.7/65.0/- & 84.4/63.8/- & 87.7/74.6/81.2 \\

GLCF~\cite{yao2023learning}  & 84.9/74.5/- & 99.4/94.2/- & 62.9/83.4/- & 57.6/80.4/- & 85.2/85.6/- & 78.0/83.6/80.8 \\
DADF~\cite{yao2023visual}  & 75.8/74.8/- & 98.7/98.4/- & 76.7/85.3/- & 66.2/88.4/- & 78.6/94.2/- & 79.2/88.2/83.7 \\

LogicAD~\cite{jin2025logicad}$\dagger$  & -/-/93.1 & -/-/81.6 & -/-/98.1 & -/-/83.8 & -/-/73.4 & -/-/86.0 \\
THFR~\cite{guo2023template}  & -/-/78.0 & -/-/97.1 & -/-/88.3 & -/-/73.7 & -/-/92.7 & -/-/86.0 \\
SINBAD~\cite{cohen2023set} & \textbf{97.7}/85.9/- & 97.1/91.7/- & 88.9/78.9/- & \textbf{81.1}/\textbf{92.4}/- & 91.5/78.3/- & 91.2/85.2/88.3 \\
GRAD~\cite{dai2024generating}  & -/-/81.2 & -/-/97.6 & -/-/99.7 & -/-/76.6 & -/-/85.4 & -/-/87.5 \\

SLSG~\cite{yang2023slsg}  & -/-/88.9 & -/-/99.1 & -/-/\textbf{95.5} & -/-/79.4 & -/-/88.5 & -/-/90.3 \\
LogSAD~\cite{zhang2025towards}$\dagger$ & -/-/\textbf{95.7} & -/-/95.2 & -/-/83.6 & -/-/83.2 & -/-/93.5 & -/-/90.2 \\
\textcolor{gray}{SAM-LAD~\cite{peng2025sam}$*$}  & \textcolor{gray}{96.7/85.2/-} & \textcolor{gray}{98.7/96.5/-} & \textcolor{gray}{97.2/79.2/-} & \textcolor{gray}{95.2/77.9/-} & \textcolor{gray}{91.4/88.6/-} & \textcolor{gray}{95.8/85.5/90.7} \\
\textcolor{gray}{PSAD~\cite{kim2024few}$*$} & \textcolor{gray}{100.0/84.9/-} & \textcolor{gray}{99.1/98.2/-} & \textcolor{gray}{100.0/89.8/-} & \textcolor{gray}{99.3/95.7/-} & \textcolor{gray}{91.9/89.3/-} & \textcolor{gray}{98.1/91.6/94.9} \\
\textcolor{gray}{CSAD~\cite{hsieh2024csad}$*$} & \textcolor{gray}{94.4/91.1/-} & \textcolor{gray}{94.9/95.6/-} & \textcolor{gray}{99.5/97.8/-} & \textcolor{gray}{99.9/93.2/-} & \textcolor{gray}{ 94.8/92.2/-} & \textcolor{gray}{96.7/94.0/95.3} \\
\midrule

\textbf{UniSLAD}$\dagger$ & 93.5/\textbf{87.2}/90.0 & 99.6/\textbf{99.7}/\textbf{99.7} & 93.9/\textbf{95.9}/94.6 & 78.5/91.8/85.8 & \textbf{92.7}/\textbf{97.2}/\textbf{95.3} & \textbf{91.6}/\textbf{94.4}/\textbf{93.1} \\
\bottomrule
\end{tabular}
\end{table*}

\textbf{MVTec LOCO AD: }
In this dataset, we report I-AUROC results separately for logical, structural, and average performance, as shown in Table~\ref{tab:auroc_mvtec_loco}. Methods that primarily target structural anomalies, such as RD4AD~\cite{deng2022anomaly} and PatchCore~\cite{9879738}, achieve strong results on structural categories but perform poorly on logical anomalies, limiting their overall effectiveness. Conversely, approaches such as GCAD~\cite{10.1007/s11263-022-01578-9} and ComAD~\cite{liu2023component} improve logical anomaly detection but exhibit substantial drops in structural anomaly performance, yielding little overall benefit. Other methods, such as DADF~\cite{yao2023visual}, yield relatively balanced results across both anomaly types; however, their performance remains at a moderate level, leaving substantial scope for further enhancement. Several additional methods show large disparities between logical and structural performance, further constraining their utility. 

In contrast, the proposed UniSLAD framework achieves a consistently balanced performance across both structural and logical anomalies, yielding an overall AUROC of 93.1\%. While recent methods like PSAD~\cite{kim2024few} and CSAD~\cite{hsieh2024csad} report higher raw metrics, they rely on additional annotations or supplementary supervisory signals. Our approach, however, sets a new benchmark by outperforming all existing methods that operate without extra annotations.

\subsection{Ablation Study}
To assess the contribution of each component in UniSLAD, we performed ablation experiments on the MVTec LOCO AD dataset. Table~\ref{tab:ablation} reports the I-AUROC results under different module configurations, including the ResNet-based Feature Extractor (RFE), the ViT-based Feature Extractor (VFE), the Mahalanobis Transform (MT), the Patch-level Memory Bank (PMB), and the Image-level Distribution Map (IDM).
 
Experiments 1, 2, and 3 compare two feature extractors. Using only RFE (Exp. 1) or only VFE (Exp. 2) results in limited performance, while combining both (Exp. 3) substantially improves anomaly detection accuracy, highlighting their complementary strengths. 
The effect of the Patch-level Memory Bank is shown by comparing Exp. 5 (without PMB) and Exp. 6 (with PMB), which demonstrates clear gains from storing representative patch features. Similarly, Exp. 7 and Exp. 8 validate the effectiveness of the Image-level Distribution Map.  
Comparing Exp. 1 (without MT) and Exp. 4 (with MT) reveals that MT significantly improves performance. This confirms the importance of decorrelating and normalizing features before similarity search.  

Experiment 9 integrates all modules and achieves the best results, 
with I-AUROC scores of 91.6\%, 94.4\%, and 93.1\% on logical, structural, and average metrics, respectively, 
demonstrating the complementary benefits of all components.

\subsection{Implications of Image Feature Converter}
The results of different Image Feature Converter (IFC) strategies are reported in Table~\ref{tab:image feature comparison}. Experiment 1, which uses simple averaging, serves as a baseline.  
Replacing mean pooling with PMP yields clear performance gains, as shown by the improvement from Experiment 1 to Experiment 3, confirming the benefit of emphasizing high-response regions.  
Furthermore, combining PMP with LUM provides additional robustness by reducing the impact of outliers, leading to further improvements from Experiment 3 to Experiment 4.  
The proposed fusion of PMP and LUM (Exp. 4) achieves the best performance, with I-AUROC scores of 91.6\% (logical), 94.4\% (structural), and 93.1\% (average), outperforming the baseline by 1.9\%, 0.3\%, and 1.4\%, respectively.

\begin{table}[t]
    \centering
    \caption{Ablation study of different modules on MVTec LOCO AD dataset. I-AUROC(\%) (logical/structural/average). 
    }   
    \label{tab:ablation} 
    \begin{tabular}{ccccccc}
    \hline

        No & RFE & VFE & MT & PMB & IDM  &{I-AUROC}  \\ \hline

        1& \checkmark & - & - & \checkmark & - &  73.4/90.3/80.5  \\
        2& - & \checkmark & - & \checkmark & -  & 73.7/76.9/74.8 \\
        3& \checkmark & \checkmark & - & \checkmark & -  & 76.0/91.0/82.5 \\

        4& \checkmark & - & \checkmark & \checkmark & - &  81.7/92.5/86.1 \\
        5& \checkmark & - & - & - & \checkmark &  91.1/86.2/88.6 \\
        6& \checkmark & - & - & \checkmark & \checkmark  & 84.7/91.9/88.7 \\

        7& - & \checkmark & \checkmark & \checkmark & -  & 79.5/80.4/79.7 \\

        8& - & \checkmark & \checkmark & \checkmark & \checkmark  &  83.6/84.0/83.7 \\

        9& \checkmark & \checkmark & \checkmark & \checkmark & \checkmark  & \textbf{91.6}/\textbf{94.4}/\textbf{93.1} \\
        \hline
    \end{tabular}
\end{table}

\begin{table}[t]

    \centering

    \caption{Comparisons of different image feature converters on MVTec LOCO AD dataset. I-AUROC(\%).}    

    \label{tab:image feature comparison}

    \begin{tabular}{p{1cm}p{3cm}p{3cm}}

    \hline

        No & IFC  &{I-AUROC}  \\ \hline
        1& mean & 89.7/94.1/91.7  \\
        2& LUM & 89.5/91,8/90.3 \\
        3& PMP & 89.7/94.3/91.9 \\                                                        
        4& LUM + PMP & \textbf{91.6}/\textbf{94.4}/\textbf{93.1} \\
        \hline
    \end{tabular}
\end{table}

\label{sec:datasets}

\vspace{-3pt}

\subsection{Case Studies}

\begin{figure}[t]
    \centering
    \includegraphics[width=0.5\textwidth]{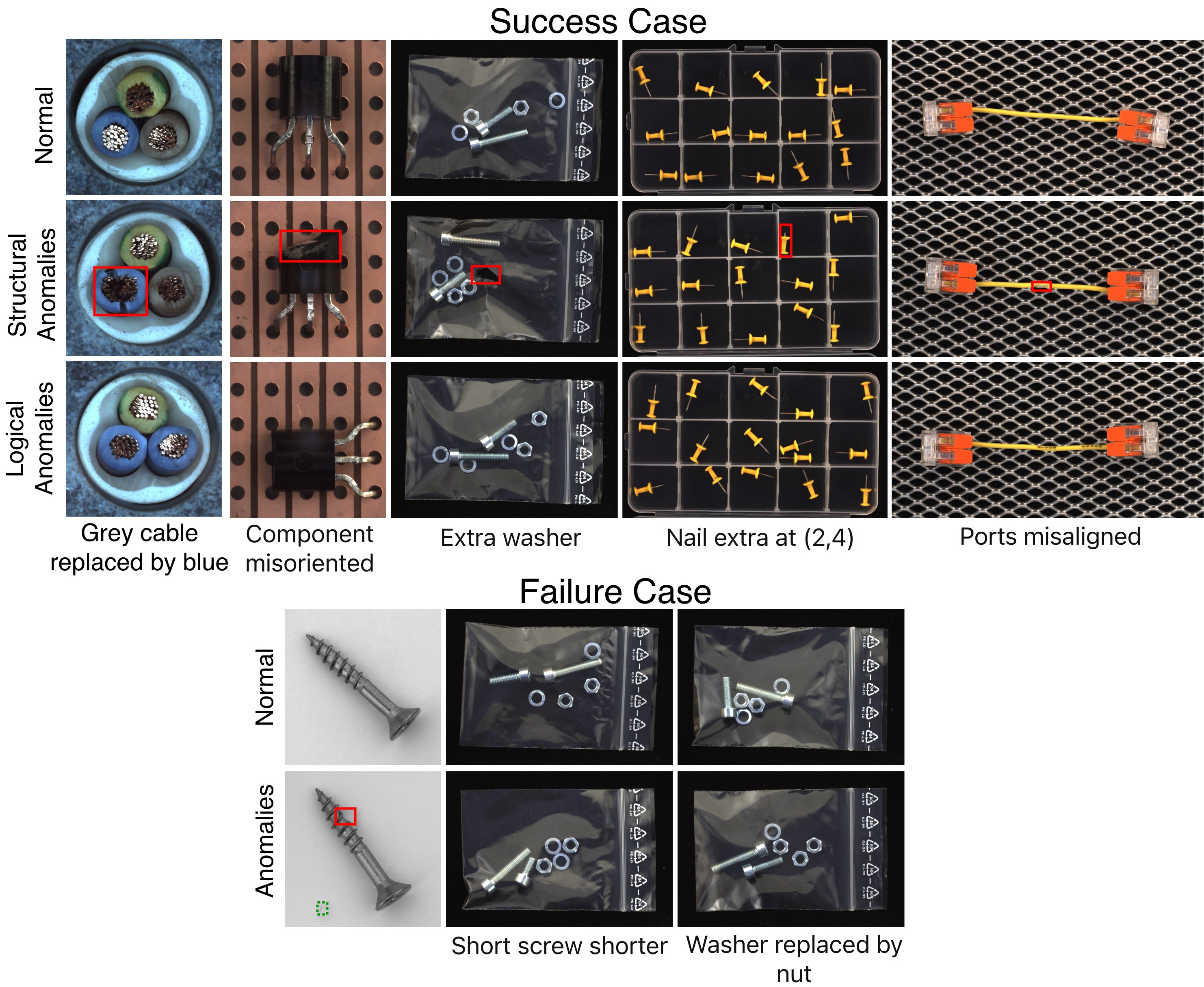} 
    \caption{Representative success and failure cases from the MVTec AD and MVTec LOCO datasets. The red solid box indicates the location of structural anomalies, and the green dashed box marks the position of background noise.}
    \label{case study}
\end{figure}

As demonstrated by the success case in Fig.~\ref{case study}, the proposed model achieves strong performance in detecting both structural and logical anomalies across various categories of the MVTec AD and MVTec LOCO AD datasets, validating its effectiveness and generalization capability.

However, for a small number of specific cases, the performance of our proposed method declined. We conducted a detailed analysis of these cases. First, a structural defect combined with heavy background noise caused the model to detect two false anomalies (Fig.~\ref{case study}, Failure Case col. 1). Second, for the MVTec LOCO Screw Bag category (cols. 2–3), our model struggled with high visual similarities between washers and nuts, alongside the subtle variations in screw lengths that define logical anomalies. Analysis of other methods revealed that methods that do not employ a segmentation module (e.g., DADF~\cite{yao2023visual} and SINBAD~\cite{cohen2023set}) exhibit lower performance in detecting logical anomalies within the Screw Bag category, whereas segmentation-based approaches such as SAM-LAD~\cite{peng2025sam} and PSAD~\cite{kim2024few} achieve substantially higher accuracy. This gap exists because feature comparison fails to capture subtle variations in the Screw Bag category, whereas segmentation yields more discriminative information at the cost of additional annotations, limiting its practical scalability.

\section{Conclusion}
In this work, we addressed the challenge of detecting both industrial structural and logical anomalies. We proposed UniSLAD, a training-free unified framework that leverages a Dual-Feature Extractor to combine the complementary strengths of CNNs and ViTs. By incorporating patch-level memory banks and image-level distribution maps, UniSLAD effectively captures anomalies across multiple granularities, enabling the detection of both fine-grained structural defects and global logical inconsistencies.
Extensive experiments on two datasets demonstrate that UniSLAD achieves competitive performance, with AUROC scores of 99.4\% and 93.1\%, respectively.  
Future work will extend the framework to real-world applications, evaluating its robustness across diverse conditions and complex environments.

\bibliographystyle{elsarticle-num}
\bibliography{refs}

@INPROCEEDINGS{9879738,
  author={Roth, Karsten and Pemula, Latha and Zepeda, Joaquin and Schölkopf, Bernhard and Brox, Thomas and Gehler, Peter},
  booktitle={Proc. IEEE/CVF Conf. Comput. Vis. Pattern Recognit.},
  title={Towards Total Recall in Industrial Anomaly Detection}, 
  year={2022},
  volume={},
  number={},
  pages={14298-14308}}

@INPROCEEDINGS{5206848,
  author={Deng, Jia and Dong, Wei and Socher, Richard and Li, Li-Jia and Kai Li and Li Fei-Fei},
  booktitle={Proc. IEEE Conf. Comput. Vis. Pattern Recognit.},
  title={ImageNet: A large-scale hierarchical image database}, 
  year={2009},
  volume={},
  number={},
  pages={248-255}}

@article{dosovitskiy2020image,
  title={An image is worth 16x16 words: Transformers for image recognition at scale},
  author={Dosovitskiy, Alexey and Beyer, Lucas and Kolesnikov, Alexander and Weissenborn, Dirk and Zhai, Xiaohua and Unterthiner, Thomas and Dehghani, Mostafa and Minderer, Matthias and Heigold, Georg and Gelly, Sylvain and others},
  journal={arXiv preprint arXiv:2010.11929},
  year={2020}
}

@article{agarwal2005geometric,
  title={Geometric approximation via coresets},
  author={Agarwal, Pankaj K and Har-Peled, Sariel and Varadarajan, Kasturi R and others},
  journal={Combinatorial and computational geometry},
  volume={52},
  number={1},
  pages={1--30},
  year={2005}
}

@INPROCEEDINGS{10647438,
  author={Sugawara, Shota and Imamura, Ryuji},
  booktitle={Proc. IEEE Int. Conf. Image Process.},
  title={PUAD: Frustratingly Simple Method for Robust Anomaly Detection}, 
  year={2024},
  volume={},
  number={},
  pages={842-848}}

@inproceedings{liu2024image,
  title={Image alignment-based patch distribution framework for anomaly detection},
  author={Liu, Yue and Ma, Ling and Jiang, Huiqin},
  booktitle={Proc. 4th Int. Conf. Comput. Vis. Data Mining},
  volume={13063},
  pages={163--172},
  year={2024},
  organization={SPIE}
}

@article{mahalanobis2018generalized,
  title={On the generalized distance in statistics},
  author={Mahalanobis, Prasanta Chandra},
  journal={Sankhy{\=a}: The Indian Journal of Statistics, Series A (2008-)},
  volume={80},
  pages={S1--S7},
  year={2018},
  publisher={JSTOR}
}

@article{10.1007/s11263-022-01578-9,
author = {Bergmann, Paul and Batzner, Kilian and Fauser, Michael and Sattlegger, David and Steger, Carsten},
title = {Beyond Dents and Scratches: Logical Constraints in Unsupervised Anomaly Detection and Localization},
year = {2022},
issue_date = {Apr 2022},
publisher = {Kluwer Academic Publishers},
address = {USA},
volume = {130},
number = {4},
journal = {Int. J. Comput. Vision},
month = apr,
pages = {947–969}
}

@INPROCEEDINGS{8954181,
  author={Bergmann, Paul and Fauser, Michael and Sattlegger, David and Steger, Carsten},
  booktitle={Proc. IEEE/CVF Conf. Comput. Vis. Pattern Recognit.}, 
  title={MVTec AD — A Comprehensive Real-World Dataset for Unsupervised Anomaly Detection}, 
  year={2019},
  volume={},
  number={},
  pages={9584-9592}}

@inproceedings{zavrtanik2021draem,
  title={Draem-a discriminatively trained reconstruction embedding for surface anomaly detection},
  author={Zavrtanik, Vitjan and Kristan, Matej and Sko{\v{c}}aj, Danijel},
  booktitle={Proc. IEEE/CVF Int. Conf. Comput. Vis.},
  pages={8330--8339},
  year={2021}
}

@inproceedings{deng2022anomaly,
  title={Anomaly detection via reverse distillation from one-class embedding},
  author={Deng, Hanqiu and Li, Xingyu},
  booktitle={Proc. IEEE/CVF Conf. Comput. Vis. Pattern Recognit.},
  pages={9737--9746},
  year={2022}
}

@article{liu2023component,
  title={Component-aware anomaly detection framework for adjustable and logical industrial visual inspection},
  author={Liu, Tongkun and Li, Bing and Du, Xiao and Jiang, Bingke and Jin, Xiao and Jin, Liuyi and Zhao, Zhuo},
  journal={Advanced Engineering Informatics},
  volume={58},
  pages={102161},
  year={2023},
  publisher={Elsevier}
}

@inproceedings{zhang2024contextual,
  title={Contextual affinity distillation for image anomaly detection},
  author={Zhang, Jie and Suganuma, Masanori and Okatani, Takayuki},
  booktitle={Proc. IEEE/CVF Winter Conf. Appl. Comput. Vis.},
  pages={149--158},
  year={2024}
}

@article{yao2023learning,
  title={Learning global-local correspondence with semantic bottleneck for logical anomaly detection},
  author={Yao, Haiming and Yu, Wenyong and Luo, Wei and Qiang, Zhenfeng and Luo, Donghao and Zhang, Xiaotian},
  journal={IEEE Transactions on Circuits and Systems for Video Technology},
  volume={34},
  number={5},
  pages={3589--3605},
  year={2023},
  publisher={IEEE}
}

@article{yao2023visual,
  title={Visual anomaly detection via dual-attention transformer and discriminative flow},
  author={Yao, Haiming and Luo, Wei and Yu, Wenyong},
  journal={arXiv preprint arXiv:2303.17882},
  year={2023}
}

@article{peng2025sam,
  title={Sam-lad: Segment anything model meets zero-shot logic anomaly detection},
  author={Peng, Yun and Lin, Xiao and Ma, Nachuan and Du, Jiayuan and Liu, Chuangwei and Liu, Chengju and Chen, Qijun},
  journal={Knowledge-Based Systems},
  volume={314},
  pages={113176},
  year={2025},
  publisher={Elsevier}
}

@inproceedings{batzner2024efficientad,
  title={Efficientad: Accurate visual anomaly detection at millisecond-level latencies},
  author={Batzner, Kilian and Heckler, Lars and K{\"o}nig, Rebecca},
  booktitle={Proc. IEEE/CVF Winter Conf. Appl. Comput. Vis.},
  pages={128--138},
  year={2024}
}

@inproceedings{jin2025logicad,
  title={Logicad: Explainable anomaly detection via vlm-based text feature extraction},
  author={Jin, Er and Feng, Qihui and Mou, Yongli and Lakemeyer, Gerhard and Decker, Stefan and Simons, Oliver and Stegmaier, Johannes},
  booktitle={Proc. AAAI Conf. Artif. Intell.},
  volume={39},
  number={4},
  pages={4129--4137},
  year={2025}
}

@inproceedings{guo2023template,
  title={Template-guided hierarchical feature restoration for anomaly detection},
  author={Guo, Hewei and Ren, Liping and Fu, Jingjing and Wang, Yuwang and Zhang, Zhizheng and Lan, Cuiling and Wang, Haoqian and Hou, Xinwen},
  booktitle={Proc. IEEE/CVF Int. Conf. Comput. Vis.},
  pages={6447--6458},
  year={2023}
}

@article{cohen2023set,
  title={Set Features for Anomaly Detection},
  author={Cohen, Niv and Tzachor, Issar and Hoshen, Yedid},
  journal={arXiv preprint arXiv:2311.14773},
  year={2023}
}

@inproceedings{dai2024generating,
  title={Generating and reweighting dense contrastive patterns for unsupervised anomaly detection},
  author={Dai, Songmin and Wu, Yifan and Li, Xiaoqiang and Xue, Xiangyang},
  booktitle={Proc. AAAI Conf. Artif. Intell.},
  volume={38},
  number={2},
  pages={1454--1462},
  year={2024}
}

@article{yang2023slsg,
  title={Slsg: Industrial image anomaly detection by learning better feature embeddings and one-class classification},
  author={Yang, Minghui and Liu, Jing and Yang, Zhiwei and Wu, Zhaoyang},
  journal={arXiv preprint arXiv:2305.00398},
  year={2023}
}

@inproceedings{zhang2025towards,
  title={Towards training-free anomaly detection with vision and language foundation models},
  author={Zhang, Jinjin and Wang, Guodong and Jin, Yizhou and Huang, Di},
  booktitle={Proceedings of the IEEE/CVF Conference on Computer Vision and Pattern Recognition},
  pages={15204--15213},
  year={2025}
}

@inproceedings{kim2024few,
  title={Few shot part segmentation reveals compositional logic for industrial anomaly detection},
  author={Kim, Soopil and An, Sion and Chikontwe, Philip and Kang, Myeongkyun and Adeli, Ehsan and Pohl, Kilian M and Park, Sang Hyun},
  booktitle={Proc. AAAI Conf. Artif. Intell.},
  volume={38},
  number={8},
  pages={8591--8599},
  year={2024}
}

@inproceedings{he2016deep,
  title={Deep residual learning for image recognition},
  author={He, Kaiming and Zhang, Xiangyu and Ren, Shaoqing and Sun, Jian},
  booktitle={Proc. IEEE/CVF Conf. Comput. Vis. Pattern Recognit.},
  pages={770--778},
  year={2016}
}

@article{hsieh2024csad,
  title={Csad: Unsupervised component segmentation for logical anomaly detection},
  author={Hsieh, Yu-Hsuan and Lai, Shang-Hong},
  journal={arXiv preprint arXiv:2408.15628},
  year={2024}
}

@inproceedings{li2025s2tkd,
  title={S2TKD: Dual-Student Knowledge Distillation for Industrial Visual Anomaly Detection and Localization},
  author={Li, Changyi and Yang, Chao and Xiao, Yu},
  booktitle={Proc. IEEE Int. Conf. Big Data},
  pages={2386--2395},
  year={2025},
  organization={IEEE}
}

\end{document}